\documentclass[letterpaper, 10 pt, conference]{ieeeconf}  

\IEEEoverridecommandlockouts                              
\usepackage[utf8]{inputenc}
\usepackage[T1]{fontenc}
\usepackage{amssymb}
\usepackage{amsmath}
\usepackage[ruled]{algorithm2e}
\usepackage{amsfonts}
\usepackage{dsfont}
\usepackage{graphicx}   
\usepackage{url}
\usepackage[backend=biber, style=ieee]{biblatex}
\usepackage{caption}
\usepackage{subcaption} 
\usepackage{stfloats}
\usepackage{multirow}
\usepackage{float}
\usepackage[dvipsnames]{xcolor}
\usepackage[bottom]{footmisc}
\floatstyle{plaintop}
\restylefloat{table}

\usepackage{xcolor,colortbl}

\definecolor{Gray}{gray}{0.85}
\definecolor{LightCyan}{rgb}{0.88,1,1}

\newcolumntype{a}{>{\columncolor{Gray}}c}

\title{\LARGE \bf
Exploiting map information for self-supervised learning \\ in motion forecasting
}

\author{Caio Azevedo$^{1, 2}$, Thomas Gilles$^{1, 3}$, Stefano Sabatini$^{1}$, Dzmitry Tsishkou$^{1}$
\thanks{$^{1}$IoV team, Paris Research Center, Huawei Technologies France}%
\thanks{$^{2}$Ecole Polytechnique, Computer Science Department}%
\thanks{$^{3}$MINES ParisTech, PSL University, Center for robotics}%
}

\begin{document}

\maketitle
\thispagestyle{empty}
\pagestyle{empty}


\begin{abstract}
    Inspired by recent developments regarding the application of self-supervised learning (SSL), we devise an auxiliary task for trajectory prediction that takes advantage of map-only information such as graph connectivity with the intent of improving map comprehension and generalization. We apply this auxiliary task through two frameworks - multitasking and pretraining. In either framework we observe significant improvement of our baseline in metrics such as $\mathrm{minFDE}_6$ (as much as 20.3\%) and $\mathrm{MissRate}_6$ (as much as 33.3\%), as well as a richer comprehension of map features demonstrated by different training configurations. The results obtained were consistent in all three data sets used for experiments: Argoverse, Interaction and NuScenes. We also submit our new pretrained model's results to the Interaction challenge and achieve \textit{1st} place with respect to $\mathrm{minFDE}_6$ and $\mathrm{minADE}_6$.
\end{abstract}

\begin{figure*}[b]
\centerline{\includegraphics[width=2.0\columnwidth]{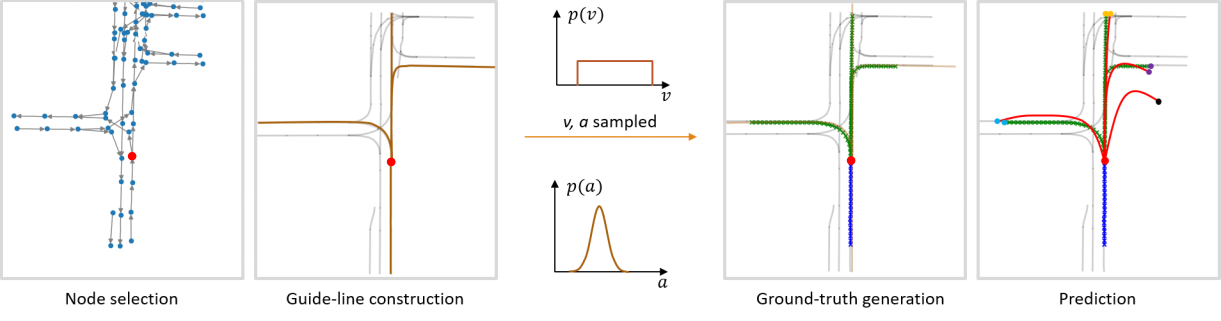}}
\caption{Design of Map Trajectories pretraining task.}
\label{fig:aux_task}
\end{figure*}

\section{INTRODUCTION}
    SSL has seen widespread application in areas such as natural language processing and computer vision, where large-scale unlabeled data sets are freely available. However, due to the lack of such large-scale data sets in motion forecasting and its difficulties in terms of data collection, only recently has SSL received attention as a means of improving performance in the prediction and planning field \cite{ssl-lanes2022}, \cite{breuer2021quovadis}. There are still, therefore, many paths to explore in regards to new auxiliary tasks, the different frameworks in which these can be coupled with the trajectory prediction task and the optimal performance that can be achieved through the tuning of the resulting new hyperparameters.
    
    We try therefore to analyze the impact of one specially designed auxiliary task on our current models to gain a deeper understanding of the effects of SSL in motion forecasting, in particular through the usage of freely available map information. More specifically, we show that in either a multitasking framework or a pretraining and then fine-tuning framework this auxiliary task is able to improve metrics. Furthermore, we remark that using map features such as graph connectivity in SSL helps the model understand better the road geometry and thus restrict the predictions to more probable paths.


\section{RELATED WORK}

\label{sec:citations}

    Learning-based methods are very effective for motion forecasting, whether they extend traditional physics-based techniques \cite{jouaber2021nnakf, song2021learning} or rely on fully data-driven pipelines \cite{altche2017lstm}. Their efficiency comes from their ease of modelling cross-data relationships, mostly for agent interaction \cite{yuan2021agentformer, jia2022multi} or map understanding \cite{liang2020learning, gilles2021home}. These relationships can be synthesized by a wide array of techniques, namely graphs \cite{salzmann2020trajectron++, mohamed2020social, khandelwal2020if}, attention \cite{mercat2020multi, messaoud2020multi, gao2020vectornet, ngiam2021scene} or other pooling methods \cite{alahi2016social, gupta2018social, ye2021tpcn}.

    However, due to hidden driver variables such as destination and driving style, the future trajectory can have multiple possibilities. Therefore trajectory prediction output is required to be multi-modal, and some methods leverage priors from either stored trajectory sets \cite{chai2020multipath, phan2020covernet} or the map graph \cite{zhang2020map, deo2022multimodal} to yield multiple proposals. Many works use the Winner-Take-All loss \cite{liang2020learning, ye2021tpcn, liu2021multimodal}, but as multi-modal supervision is impossible, this loss is inefficient and can only train one modality at a time. 
    More recently, many works have tried to improved their multi-modal prediction by using model ensembles for  clustering \cite{varadarajan2022multipath++, nayakanti2022wayformer} or for transfer learning \cite{ye2022dcms}, but these methods remain very expensive training time-wise, as they require to train N times more models, and in some case to infer all of them. 

    Pretraining, and more specifically self-supervision, has been prevalent in many widely explored learning fields such as NLP \cite{devlin2018bert} and vision \cite{dosovitskiy2020image, he2022masked}. In other tasks of autonomous driving such as control, some methods design synthetic ground truths \cite{chekroun2021gri} in order to help the learning of their driving agents. Concurrent to this work, PreTraM \cite{xu2022pretram} applies contrastive learning between local map rasters and trajectories to reinforce the learned relationship between both, while SSL-Lanes \cite{ssl-lanes2022} uses a multitasking framework to analyze the performance improvement of four auxiliary tasks.


\section{METHOD}
\label{sec:method}

        We proceed as following. First, we describe our base prediction model, on top of which we will apply the auxiliary task in different frameworks. We continue with the description of each element of this task, and finally detail its application in each of the frameworks, either pretraining or multitask.

    \subsection{Base trajectory prediction model}
    
        Our encoder is the same as in the GOHOME model \cite{gilles2021gohome}, which uses attention and graph convolutions to update agent features with map information and to model agent interactions. However, instead of decoding a heatmap of the probability density of the final position of the tracked agent, we like most of the state-of-the-art decode the $k$ full trajectory predictions directly through a multi-layer perceptron and the probability logits of each prediction separately through a simple linear layer.
        
        The trajectory loss function in this case selects the closest prediction to the ground-truth by the usual winner-takes-all method according to $\mathrm{minFDE}_k$. Once the closest prediction is selected, we apply a smooth L1 loss \cite{girshick2015fastrcnn} between the $N$ points in it and the $N$ points in the ground-truth, and finally average the results to get the main trajectory loss $\mathcal{L}_{traj}$.
        
        There is also, however, a loss associated to the prediction probabilities $\mathcal{L}_{prob}$ which in our case is the same as the one used in TPCN \cite{ye2021tpcn}. In the end we combine these losses to have $\mathcal{L}_{main} = \mathcal{L}_{traj} + \mathcal{L}_{prob}$.

    \subsection{Design of the \textbf{Map Trajectories} auxiliary task}
    
        Our aim is to design an auxiliary task that uses map information as a way of improving map comprehension. However, such a task needs to share important features with the main goal of trajectory prediction in order to avoid forgetting the information learned \cite{mccloskey1989} or having conflicting information from both tasks. One essential feature of motion forecasting is multi-modality, and a natural approach to incorporate it to the SSL task while exploiting the map is to make the network predict, given a starting position, all trajectories that an agent in that position could take in a given time horizon. Each trajectory can be split into two parts - the past and the future. All trajectories generated for the same starting position share the same past, but their possible futures vary.
    
        \subsubsection{Map exploration}
    
            Each data set provides its HD-map as a graph that consists of lanelets (nodes) and edges. The lanelets represent sections of the roads 10 to 20 meters long on average. They contain the succession of coordinates of the center line of its associated lane segment, also giving its direction. The edges of the graph represent the connectivity of the lanelets. For clarity, in the following we differentiate a "path" -- a sequence of lanelets that may be travelled by an agent, from a "trajectory" -- a precise sequence of coordinates taken in regular time intervals and of definite length.
            
            The pretraining data consists of a list of such lanelets along with the local graph each is a part of, allowing the easy choice of a starting position from which the all the possible paths will be built using the connectivity of the graph every time a sample is taken.
        
            From the starting lanelet, we do a depth-first search to find all possible paths. We stop the search once a maximum distance is reached, or when there are no more successor nodes. We store the paths found as lists of lanelet IDs and for each of these lists we concatenate their respective center lines to create a "guide-line" for the possible trajectories.
    
        \subsubsection{Synthetic speeds and accelerations}
        
            In order to bring the network input as close as possible to that of the main trajectory prediction task, we create from sampled values of speed and acceleration the ground-truths of the possible trajectories and a synthetic past history.
            
            We sample the agent's initial speed from a uniform distribution, which allows the model to adapt to the different speed distributions found in each data set more easily. In a certain percentage of samples, we also take a random acceleration from a Laplace distribution, centered on 0 and with a scale chosen to fit the empirical data. This acceleration is kept constant throughout the past, but to create diversity in the future modalities, we add a random noise to the acceleration in the future part of the trajectory, also taken from a Laplace distribution of smaller scale.
        
        \subsubsection{Ground-truth and past generation}
        
            Using the guide-line generated by the node paths, we interpolate its coordinates to create a ground-truth trajectory that has constant acceleration, consistent with the speed and acceleration sampled. We take care to extrapolate the trajectory if there are not enough points or to cut it off earlier whenever the desired number of points is achieved, so that every possible trajectory is an array of equal size. The past is generated in the same way, except that we simply traverse the starting lanelet's predecessors instead of its successors for the node ID list, and only generate one trajectory.
            
            Finally, to simulate perception noise encountered in the real data, we add an independent Gaussian noise to each step of the past trajectory.
        
        \subsubsection{Network output}
        
            The network outputs $n_{pred}$ predictions of map trajectories. The loss function is based upon the matching of predictions and ground-truths.
            
            Most often, the number of actual ground-truths $n_{gt}$ is less than $n_{pred}$, in which case some predictions are left unmatched to any ground-truth. We run then into an assignment problem of which predictions should be matched to a certain ground-truth for the loss calculation. To solve this, we take, using the Hungarian algorithm \cite{kuhn1955hungarian}, the set of predictions that minimizes the mean L2 distance of its points to the points in the ground-truths. This produces a set of matched ground-truths $m_1,...,m_{n_{gt}} \in \{ 1, ..., n_{pred} \}$, $m_i \neq m_j$ if $i \neq j$. Let $N$ be the number of points in each trajectory, $p_{ij}$ be point $j$ of prediction $i$ and $q_{ij}$ be point $j$ of ground-truth $i$. The loss is then taken as:
            \[ \mathcal{L}_{aux}(i, j) =
            \begin{cases}
                \frac{1}{N} \sum_{k = 1}^N \| p_{ik} - q_{jk} \| & \text{if } m_j = i, \\
                0 & \text{otherwise.}
            \end{cases}
             \]
    
            In order to fully make use of our model's architecture, we also decode probabilities associated to each prediction from the map-aware agent encoding. However, since in this case we have multiple ground-truths, the probability loss of each prediction is calculated based on the closest ground-truth.
    
    \subsection{Application of auxiliary task}
    
        We compare the performance of applying the Map Trajectories task in two distinct frameworks.
        
        \subsubsection{Pretraining framework (PT)}
        
            We use the same architecture of our base trajectory prediction model to instead perform the auxiliary task. We then load the resulting pretrained weights for fine-tuning in motion forecasting.
        
            \begin{figure}[h]
            \centerline{\includegraphics[width=1\columnwidth]{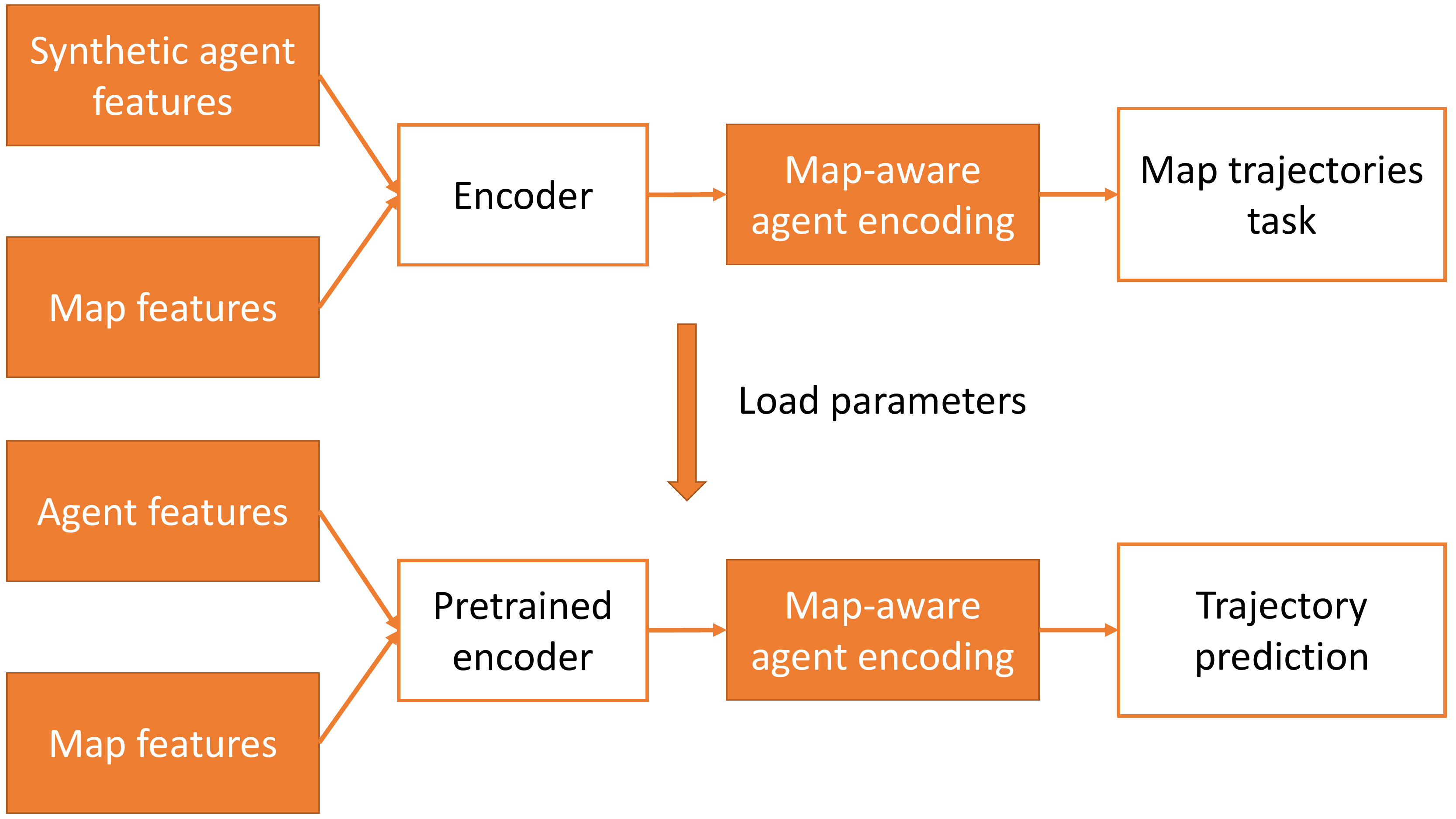}}
            \caption{Pretraining pipeline.}
            \label{fig:pt_pipeline}
            \end{figure}
            
            The pretraining task is done on all maps available.
        
        \subsubsection{Multitasking framework (MT)}
        
            The Map Trajectories task changes when applied in MT. Instead of picking random positions in the map graphs, we use the same starting position and past history of the tracked agent in the current sample, since the input has to be the same for both tasks. We then generate the possible ground-truth trajectories from this starting position, using zero acceleration and a speed that enforces an equal traversed distance to the trajectory prediction ground-truth.
            
            With this in mind, the map-aware agent encoding is decoded by one head for each of the parallel tasks. The total loss thus becomes a combination of the previous two losses:
            
            \[ \mathcal{L} = \mathcal{L}_{main} + \lambda \mathcal{L}_{aux}, \]
            
            \noindent giving an additional hyperparameter $\lambda$ that can be tuned to optimize performance.
            
            \begin{figure}[h]
            \centerline{\includegraphics[width=1\columnwidth]{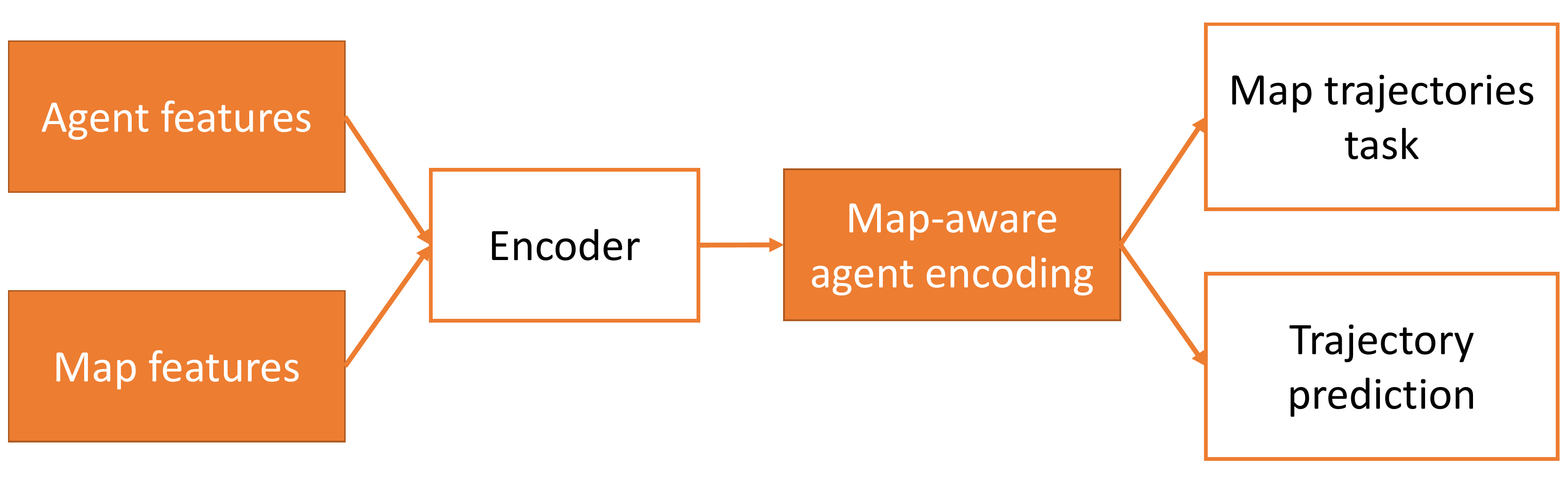}}
            \caption{Multitasking pipeline.}
            \label{fig:mt_pipeline}
            \end{figure}


\section{Experimental Results}
\label{sec:result}

    \subsection{Experimental settings}
    
        \subsubsection{Data sets}
        
            The Argoverse data set \cite{chang2019argoverse} is made of 205942 training samples, 39472 validation samples and 78143 test samples. Past histories have 2 seconds while future trajectories have 3 seconds, all sampled at 10Hz. The Interaction data set is made of 447626 training samples and 130403 validation samples, with 1 seconds histories and 3 seconds future trajectories, also sampled at 10Hz. One of its particularities is that it makes available maps differentiated by type -- which allows for generalization experiments more easily -- some of which do not have agent data in them. Finally, the NuScenes data set \cite{caesar2020nuscenes} is composed of 32186 training samples and 9041 validation samples, with 2 seconds history and 6 seconds future sampled at 2Hz.
            
        \subsubsection{Metrics}
        
            The usual metrics adopted in Argoverse and Interaction are $\mathrm{minFDE}_k$, minimal Final Displacement Error; and $\mathrm{MR}_k$, Miss Rate - both for the top $k$ predictions. We report these values for $k=6$ in Argoverse, Interaction and NuScenes.
        
        \subsubsection{Implementation details}
        
            We describe in Tab \ref{tab:distrib_parameters} the probability distribution parameters used to sample the speed, acceleration and noise value that generate the Map Trajectories during pretraining.$\mathcal{U}$ refers to the uniform distribution, $\mathcal{L}$ to the Laplace law and $\mathcal{N}$ to the normal distribution.

        \begin{table}[h]
        \caption{Pretraining Parameters}
            \begin{center}
            \begin{tabular}{c|c}
                \hline
                  Parameter & Probability Law \\
                \hline
                Speed (m/s) & $\mathcal{U}$(0, 20) \\
                Past acc (m/s$^2$)& $\mathcal{L}$(0, 1.4) \\
                Fut acc (m/s$^2$) & Past acc + $\mathcal{L}$(0, 0.9) \\
                Past noise (m) & $\mathcal{N}$(0, 1.0) \\
                \hline
            \end{tabular}
            \end{center}
            \label{tab:distrib_parameters}
        
        \end{table}
        
            All trainings have 32 epochs, be it the base trajectory prediction model, the pretraining task using the same architecture or the multitasking framework. We use in all of them a batch size of 64, with AdamW optimizer and cosine annealing learning rate schedule, starting from $3\mathrm{e}^{-4}$ and with $T_{max} = 32$.
            
            When testing the multitasking framework, we remark that different data sets require different values of $\lambda$ to compensate, for example, for more noisy trajectories such as in Argoverse, where the main loss is much higher than in Interaction and thus requires a higher weight for the auxiliary task to achieve comparable improvements. Optimal results were achieved using $\lambda = 0.1$ in Interaction, $\lambda = 0.5$ in Argoverse.

    \subsection{General improvement}
    
        In either framework, we observe an improvement of our base model, as shown in Tab. \ref{tab:general_improvement}. In this table, Mdl. represents the model used: B stands for the base model, while SP stands for "self-pretrained" and refers to loading the baseline weights and parameters into the model again for another training -- thus not using the auxiliary task at all, for reference with an equal total number of training steps. MT and PT refer to using the two SSL frameworks discussed.

        \begin{table}[H]
        \begin{minipage}{\linewidth}
        \caption{Improvement from PT (Pretraining) and MT (Multitasking) in each data set. Metrics from validation.}
            \begin{center}
            \begin{tabular}{l|l|c|c}
                \hline
                Data set & Mdl. & $\mathrm{minFDE}_6$ & $\mathrm{MR}_6$   \\
                \hline
                \multirow{4}{*}{Argoverse}  & B & 1.090 & 10.2 \\
                                            & SP   & 1.038 & 9.3 \\
                                            & MT     & 1.068 & 9.5 \\
                                            & PT & \textbf{1.031} \textcolor{Green}{(-5.4\%)} & \textbf{9.1} \textcolor{Green}{(-10.8\%)} \\
                \hline
                \multirow{4}{*}{Interaction} & B & 0.272 & 0.45  \\
                                             & SP & 0.260 & 0.44   \\
                                             & MT & 0.254  & 0.45\\
                                             & PT & \textbf{0.240} \textcolor{Green}{(-11.8\%)} & \textbf{0.30} \textcolor{Green}{(-33.3\%)} \\
                \hline
                            & B & 3.15 &  58.9  \\
                NuScenes    & SP & 2.78 &  41.4  \\
                            & PT & \textbf{2.51} \textcolor{Green}{(-20.3\%)} & \textbf{45.2} \textcolor{Green}{(-23.3\%)}\\
                            
                \hline
            \end{tabular}
            \end{center}
            \label{tab:general_improvement}
        \end{minipage}
        \end{table}

        Surprisingly, we find that the pretraining framework shows a larger improvement, contrary to established consensus on SSL in graph neural networks \cite{jin2020sslongraphs} -- indicating that at the very least the best framework highly depends on the auxiliary task itself and its implementation.
        
        We also report the results of submitting the pretrained model to the Interaction leaderboard \cite{interpret} for comparison with the state-of-the-art, as shown in Tab. \ref{tab:interaction_leaderboard} where we report other published methods on the leaderboard.
        
        \begin{table}[H]
        \begin{minipage}{\linewidth}
        \caption{Interaction leaderboard \cite{interpret}}
            \begin{center}
            \begin{tabular}{c|c| c|c}
                \hline
                 & $\mathrm{minADE}_6$ & $\mathrm{minFDE}_6$ & $\mathrm{MR}_6$ \\
                \hline
                DenseTNT \cite{gu2021densetnt} & 0.434 &  0.795      &  6.0 \\
                GOHOME \cite{gilles2021gohome} & 0.201 & 0.599       & \textbf{4.9}  \\ 
                MMTransformer \cite{huang2022multi}& 0.213 &  0.551   & 5.1  \\
                HDGT \cite{jia2022hdgt} & 0.168  &  0.478    &  5.6 \\
                \hline
                Ours (Pretrained) & \textbf{0.153} & \textbf{0.443} & 7.0  \\
                \hline
            \end{tabular}
            \end{center}
            \label{tab:interaction_leaderboard}
        
        \end{minipage}
        \end{table}
        
        We observe that with the pretrained model we achieve $1^{st}$ place both in $\mathrm{minADE}_6$ and $\mathrm{minFDE}_6$.
        
    \subsection{Richer map comprehension}
    
        In order to assess whether or not the network has achieved greater comprehension of map features such as the geometry and connectivity of the roads thanks to the pretraining, we perform a few experiments using different training configurations in the Interaction data set, which allows for the selection of maps based on their type (roundabouts, intersections, merging) for trainings.
        
        In the first training configuration, we perform a normal training that sees all the maps available, for the baseline ("Base") and for the pretrained model in which the pretraining also saw all maps ("Pretrained 1"). In a second configuration, we train on all the maps except on roundabouts with Base, Pretrained 1 and a pretrained model in which the pretraining did not see roundabouts either ("Pretrained 2").
        
        We report $\mathrm{minFDE}_6$ obtained per map type in Tab. \ref{tab:interaction_experiments}. As expected, the training that did not see roundabouts performs much worse in these than in other map types. However, both pretraining configurations show considerable improvement on roundabouts, even the one that did not see roundabouts either. From this and the fact that their metrics are comparable between each other we may conclude that for fine-tuning performance the types of maps that were seen in pretraining do not have as much impact as expected. We hypothesize that this is because from the task itself what the network is learning is how to use lanes and their connectivity in a general way.
        
\begin{figure*}[t]
        \centerline{\includegraphics[width=2.0\columnwidth]{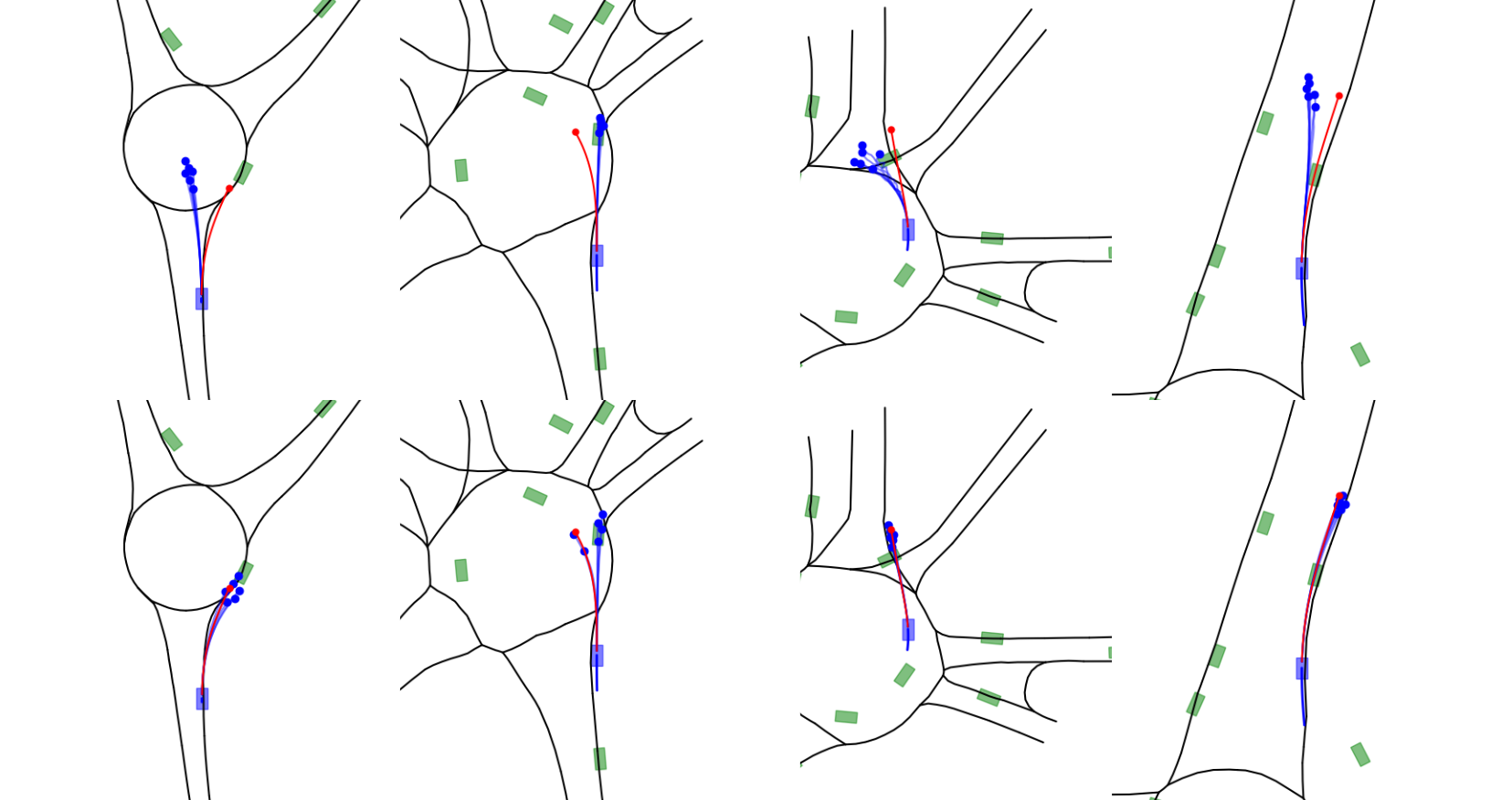}}
        \caption{Performance improvement in trajectory prediction between "Base" (top row) and "Pretrained 2" (bottom row) models when trainings did not see roundabouts. Center lines of lanes are in black. The blue rectangle is the tracked agent, while green ones are other agents present in the scene. Blue trajectories with final dots are predictions, while the red trajectory is the ground-truth.}
        \label{fig:qualitative}
        \end{figure*}

        We show qualitative examples of the pretraining improvement in Fig. \ref{fig:qualitative}. The "Base" model (top row), which has not seen any roundabout map, fails to understand the map topology and predicts the agent to either go straight or miss the exit. On the other hand, even without using the missing roundabout maps, the pretraining enables the model to have a better use of the map, and the prediction follows the roundabout curvature as well as predicts multiple modalities for the possible exits.
        
        \begin{table}[H]
        \begin{minipage}{\linewidth}
        \caption{Improvement of pretraining on Interaction by map type.}
            \begin{center}
            \begin{tabular}{c|c|c|c|c}
                \hline
                Training & \multirow{2}{*}{Model} & \multicolumn{3}{c}{$\mathrm{minFDE}_6$}   \\
                configuration & & Rndbts. & Inters. & Merg. \\
                \hline
                \multirow{2}{*}{All}& Base       & 0.345 & 0.316 & 0.153  \\
                                    & Pretr. 1 & \textbf{0.301} \textcolor{Green}{(-12.8\%)} & \textbf{0.288} & \textbf{0.137} \\
                \hline
                \multirow{3}{*}{No rndbts.} & Base & 1.117 & 0.376 & 0.178 \\
                            & Pretr. 1  & 0.806 \textcolor{Green}{(-27.8\%)} & 0.323 & \textbf{0.161} \\
                            & Pretr. 2  & \textbf{0.790} \textcolor{Green}{(-29.3\%)} & \textbf{0.322} & 0.171 \\
                \hline
            \end{tabular}
            \end{center}
            \label{tab:interaction_experiments}
        
        \end{minipage}
        \end{table}

    \subsection{Pretraining Task ablation study}
    
        In order to analyze the impact of each feature of the Map Trajectories task, we analyze the pretraining effect when removing the differents steps in trajectory generation . We first report the performance of a model trained only on fixed length paths, without varying speed ($v=10)$. We then compare pretraining with constant speed valuers without any acceleration ($PastAcc = FutAcc=0$), to the task with constant acceleration without any acceleration noise between the past and future ($ FutAcc = PasrAcc$).
        The results of each fine-tuning is shown in Tab. \ref{tab:abl}, where we also report the raw performance of the pretrained model evaluated on the validation set without any finetning.
        
        \begin{table}[H]
        \begin{minipage}{\linewidth}
        \caption{Ablation of generation steps in \textbf{Map Trajectories} on the NuScenes validation set.}
            \begin{center}
            \begin{tabular}{c|c|c |c|c}
                \hline
                Task version & Pretrained & \multicolumn{3}{c}{Finetuned} \\
                 & $\mathrm{mFDE}_6$ & $\mathrm{mADE}_5$ & $\mathrm{mFDE}_6$ & $\mathrm{MR}_6$ \\
                \hline
                Fixed length & 15.1 & 1.71 & 2.57       & 47  \\
                Fixed speed & 8.01 & 1.71 & 2.47       & 42  \\
                Fixed acc & 5.96 & 1.51  & 2.68       & 45  \\
                No past noise & 5.26 & 1.45 & 2.54       & \textbf{44}  \\
                Final & \textbf{4.79} & \textbf{1.43} & \textbf{2.51}       & 45  \\
                \hline
            \end{tabular}
            \end{center}
            \label{tab:abl}
        
        \end{minipage}
        \end{table}
        
        We observe that each generation step, by allowing more realism to real data and more diversity, improves both direct untuned performance and final finetuned metrics. Notably, we notice that raw performance correlates quite directly with the final tuned one.
        It can also be observed that the past noise improvement is relatively limited, however on other datasets such as Argoverse which has much more perception noise, we have noted this step to be crucial for good performance.
        


\section{Conclusion}
\label{sec:conclusion}

    In this paper we propose an auxiliary task that, through the use of map information such as lane connectivity, is able to improve our base models and make them understand more deeply how to navigate the map, giving additional evidence for the efficacy of applying SSL in motion forecasting. We also find that, for this task, the pretraining framework shows larger improvement than a multitasking one, indicating that a choice of framework needs to be applied on a case-by-case basis. In future work new auxiliary tasks can be devised to further enhance the performance of current state-of-the-art models.



\clearpage
\printbibliography

\end{document}